\documentclass{article}

\usepackage{arxiv}

\usepackage[utf8]{inputenc} 
\usepackage[T1]{fontenc}    
\usepackage{hyperref}       
\usepackage{url}            
\usepackage{booktabs}       
\usepackage{amsfonts}       
\usepackage{nicefrac}       
\usepackage{microtype}      
\usepackage{graphicx}\graphicspath{{images/}}
\usepackage{todonotes}
\usepackage{xcolor}
\usepackage{makecell}
\usepackage{amsmath}

\usepackage{natbib}
\usepackage{doi}

\title{On Accelerating Edge AI: Optimizing Resource-Constrained Environments}

\renewcommand{\shorttitle}{Optimizing Edge AI}

\author{
  Jacob Sander \\
  Department of Computer Science \\
  University of West Florida \\
  Pensacola, FL, USA \\
  \texttt{jhs39@students.uwf.edu} \\
  \And
  Achraf Cohen \\
  Department of Mathematics and Statistics \\
  University of West Florida \\
  Pensacola, FL, USA \\
  \texttt{acohen@uwf.edu} \\
  \And
  Venkat R. Dasari \\
  DEVCOM Army Research Laboratory \\
  Aberdeen Proving Ground, MD, USA \\
  \texttt{venkateswara.r.dasari.civ@army.mil} \\
  \And
  Brent Venable \\
  Department of Intelligent Systems and Robotics \\
  University of West Florida \\
  Institute for Human \& Machine Cognition \\
  Pensacola, FL, USA \\
  \texttt{bvenable@uwf.edu} \\
  \And
  \href{mailto:bjalaian@uwf.edu}{Brian Jalaian*} \\
  Department of Computer Science \\
  Department of Intelligent Systems and Robotics \\
  University of West Florida \\
  Institute for Human \& Machine Cognition \\
  Pensacola, FL, USA \\
  \texttt{bjalaian@uwf.edu}
}

\hypersetup{
  pdftitle={On Accelerating Edge AI: Optimizing Resource-Constrained Environments},
  pdfsubject={cs.AI},
  pdfauthor={Jacob Sander, Achraf Cohen, Venkat R. Dasari, Brent Venable, Brian Jalaian},
  pdfkeywords={Edge AI, Optimization, Resource-Constrained Environments, Neural Architecture Search, Model Compression}
}

\begin{document}

\maketitle

\begin{abstract}
Resource-constrained edge deployments demand AI solutions that balance high performance with stringent compute, memory, and energy limitations. In this survey, we present a comprehensive overview of the primary strategies for accelerating deep learning models under such constraints. First, we examine \emph{model compression} techniques-pruning, quantization, tensor decomposition, and knowledge distillation-that streamline large models into smaller, faster, and more efficient variants. Next, we explore \emph{Neural Architecture Search (NAS)}, a class of automated methods that discover architectures inherently optimized for particular tasks and hardware budgets. We then discuss \emph{compiler and deployment frameworks}, such as TVM, TensorRT, and OpenVINO, which provide hardware-tailored optimizations at inference time. By integrating these three pillars into unified pipelines, practitioners can achieve multi-objective goals, including latency reduction, memory savings, and energy efficiency-all while maintaining competitive accuracy. We also highlight emerging frontiers in \emph{hierarchical NAS}, \emph{neurosymbolic approaches}, and \emph{advanced distillation} tailored to large language models, underscoring open challenges like pre-training pruning for massive networks. Our survey offers practical insights, identifies current research gaps, and outlines promising directions for building scalable, platform-independent frameworks to accelerate deep learning models at the edge.
\end{abstract}



\section{Introduction}
Integrating Artificial Intelligence (AI) models into tactical edge computing systems is fundamentally transforming military operations and evolving battlefield technology over time. In practical applications of tactical edge computing, AI models are deployed on constrained platforms that have significant limitations in processing power, memory, and communication capabilities. These constraints can hinder the use of complex AI models for real-time tasks such as surveillance, enemy identification, predictive simulations, and navigation.

The goal is to develop intelligent and adaptive optimization strategies that can overcome these resource limitations while enhancing the performance of AI models without significantly compromising their accuracy. This is particularly critical in military contexts, where timely and precise decision-making can be pivotal to the success of a mission. Therefore, the ability to effectively utilize AI models in tactical edge computing environments is crucial for military applications. This capability allows for the creation of intelligent systems and machines that can operate in harsh conditions with limited network access, ultimately providing invaluable support to troops in the field.

\begin{figure}[h]
    \centering
    \includegraphics[width=250pt]{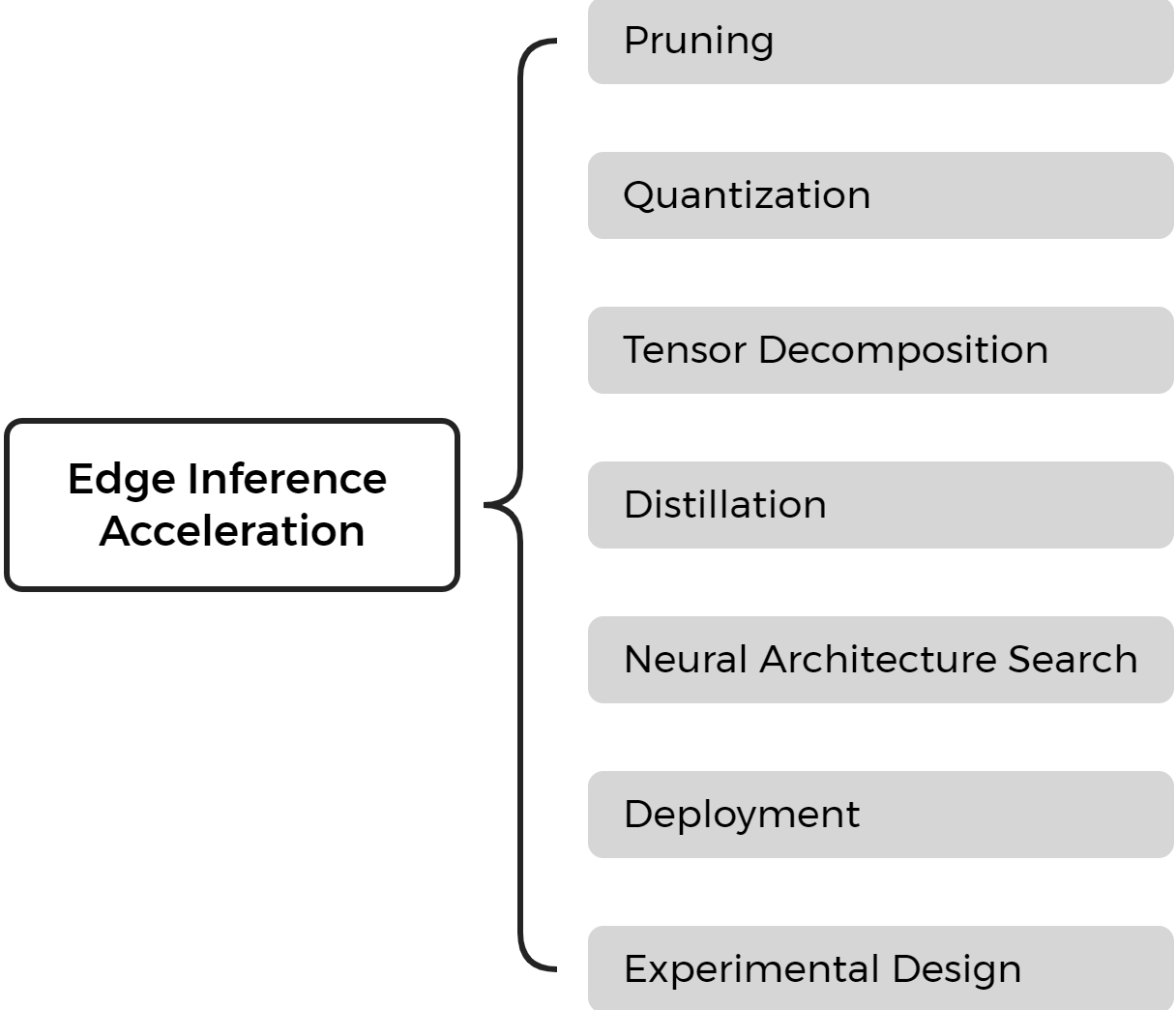}
    \caption{Strategies and Tools for Enhancing AI Edge Computing}
    \label{fig1}
\end{figure}

Our survey aims to identify emerging approaches for optimizing Artificial Intelligence (AI) models to accelerate model inference. We focus on strategies that meet both model-specific and platform-specific constraints while maintaining accuracy. The survey will examine techniques such as model pruning, quantization, knowledge distillation, neural architecture search, and hardware-aware design. Fig. \ref{fig1} presents a map of strategies and tools for edge computing acceleration. By exploring the latest advancements in AI model optimization, we hope to uncover promising methods that address the challenges of deploying AI models in resource-constrained environments. Ultimately, our goal is to contribute to developing more efficient, effective, and reliable AI-powered systems for critical applications.


We categorize optimization strategies into two main families, although in practice, scientists and engineers often combine approaches to enhance performance. First, we will examine \emph{Model Compression} techniques, which focus on transforming a trained model into a smaller version that still maintains high performance. However, Model Compression techniques lack performance guarantees, leading researchers to seek higher-performing and more innovative architectures. Consequently, we will also address \emph{Neural Architecture Search} (NAS) and Hyper-Parameter Optimization (HPO) methods that explore the configuration space to identify more efficient architectures based on defined objectives. Although these methods are formally separate from the aforementioned optimizations, we will discuss compile-time techniques that reduce model inference time during deployment. Finally, we will provide a detailed overview of experiments involving composite methods, which integrate Model Compression techniques with customized objectives and constraints to explore the NAS and HPO space.

\textcolor{black}{This paper is organized as follows: Section 2 introduces model compression techniques and their characteristics, such as quantization, pruning, and knowledge distillation; Section 3 presents neural architecture search;  
Section 4 discusses the deployment aspects related to compilers and hardware. Section 5 discusses case studies and integrated approaches.  
Finally, in Section 6, conclusions and some possible research directions are presented.}

\section{Model Compression Techniques}\label{sec:model-compression}

AI models, especially deep learning ones, tend to be over-parameterized. This over-parameterization is crucial, making model compression feasible and highly effective. Model compression refers to reducing the size and computational complexity while preserving performance (e.g., accuracy).

In tactical edge computing environments, the size and computational demands of AI models can hinder their deployment on resource-constrained platforms. Model compression techniques aim to reduce the size of larger AI models and lower computational requirements during inference, all without significantly compromising performance. Compressing a model enables the use of advanced AI capabilities in tactical edge computing environments with limited computational power, memory, and storage.

\subsection{Quantization}
\textcolor{black}{Quantization is the process of reducing the precision of weights and activations, often stored as 32-bit floating-point values, to lower bit representations such as 8-bit or binary, to optimize memory and computational efficiency \cite{jacob2018quantization}.}
Recent research on quantization in large language models (LLMs) highlights its importance for reducing model size and improving efficiency. Quantization techniques include post-training quantization (PTQ) and quantization-aware training (QAT) \cite{chen2024efficientqatefficientquantizationawaretraining}, with algorithms like LLM-QAT and SmoothQuant addressing challenges such as outliers and activation quantization \cite{wang2024art}.
\textcolor{black}{SmoothQuant addresses outliers in activation values by redistributing them to reduce their impact during quantization, enabling stable and efficient low-bit representations.}
\textcolor{black}{Post-training quantization (PTQ) modifies the model’s parameters after training without additional data or computational overhead. In contrast, quantization-aware training (QAT) incorporates quantization during training, allowing the model to adapt to the reduced precision, often yielding better performance.}
\textcolor{black}{LLM-QAT is specifically tailored for large language models, adapting quantization-aware training techniques to handle the unique challenges posed by their massive parameter scales and diverse activation ranges.}
\textcolor{black}{CVXQ leverages convex optimization techniques to enable highly flexible and efficient post-training quantization, particularly for extremely large models with billions of parameters\cite{young2024foundations}.}
Activation-Aware Weight Quantization (AWQ) \cite{lin_awq_2024}, EfficientQAT \cite{chen2024efficientqatefficientquantizationawaretraining}, and QLoRA \cite{dettmers2023qloraefficientfinetuningquantized} all present alternatives to training a full-precision model from scratch to implement quantization-aware models.
\textcolor{black}{For example, AWQ\cite{lin_awq_2024} introduces a novel mechanism to optimize weight quantization based on activation patterns, ensuring high accuracy even under aggressive quantization regimes.}
Comparative studies reveal that the optimal quantization format (integer or floating-point) varies across layers, leading to the proposal of Mixture of Formats Quantization (MoFQ) for improved performance in both weight-only and weight-activation scenarios \cite{Zhang2024}.
\textcolor{black}{The RPTQ employs a channel-wise rearrangement and clustering strategy to manage activation range variations, enabling robust 3-bit activation quantization for large language models\cite{rtpq2023}}.
\textcolor{black}{Outlier Suppression+ innovates by introducing channel-wise shifting and scaling, effectively addressing asymmetric outliers and achieving near-floating-point performance for both small and large models\cite{Wei2023}.}
These techniques can significantly reduce memory consumption, with OPT-175B (Meta's Open Pre-trained Transformer) quantization potentially leading to an 80\% reduction \cite{rtpq2023}.

Mixed precision training has emerged as an effective technique for improving the efficiency of large-scale AI models. Researchers introduced a method using half-precision floating-point numbers for weights, activations, and gradients, while maintaining a single-precision copy of weights to accumulate gradients \cite{Micikevicius2017MixedPT}. Li et al. \cite{Li2023} proposed a layered mixed-precision approach, adjusting training precisions for each layer based on its contribution to the training effect. The Channel-Wise Mixed-Precision Quantization (CMPQ) was developed \cite{ChenMXT2024}, allocating quantization precision in a channel-wise pattern for large language models (OPT-2.7B and OPT-6.7B). Other research explored mixed precision low-bit quantization for neural network language models in speech recognition \cite{Xu2021}, using techniques such as KL-divergence, Hessian trace weighted quantization perturbation, and mixed precision neural architecture search. These methods have shown notable enhancements in training speed, memory efficiency, and model compression, all while preserving performance in various deep learning applications.

Applying weight and activation quantization together improves inference speed and model memory metrics, especially for quantization-aware training \cite{gaurav_efficiency_survey, google_model_optimization}. See Table \ref{tab:quantization table}.

\begin{table}[!ht]
\footnotesize
    \centering
       \setlength{\tabcolsep}{1.5pt}
    \begin{tabular}{lllllllll}
        \hline
        \makecell{Model} & \makecell{Accuracy} & \makecell{Accuracy} & \makecell{Accuracy} & \makecell{Latency} & \makecell{Latency} & \makecell{Latency} & \makecell{Original} & \makecell{Optimized} \\ 
         
         \makecell{} & \makecell{(Original)} & \makecell{(PTQ)} & \makecell{(QAT)} & \makecell{(Original)} & \makecell{(PTQ)} & \makecell{(QAT)} & \makecell{Size (MB)} & \makecell{Size (MB)} \\ 
        \hline
        Mobilenet-v1-1-224 & .709 & .657 & .70 & 124 & 112 & 64 & 16.9 & 4.3 \\ \hline
        Mobilenet-v2-1-224 & .719 & .637 & .709 & 89 & 98 & 54 & 14 & 3.6 \\ \hline
        Inception v3 & .78 & .772 & .775 & 1130 & 845 & 543 & 95.7 & 23.9\\ \hline
        Resnet v2 101 & .770 & .768 & N/A & 3973 & 2868 & N/A & 178.3 & 44.9 \\ \hline
    \end{tabular}
    \vspace{5pt}
    \caption{Comparison of original, post-training quantized (PTQ), and quantization-aware training (QAT) in terms of accuracy, latency (ms), and size as reported in \cite{google_model_optimization}.}
    \label{tab:quantization table}
\end{table}

In certain situations, hardware memory constraints can affect inference performance. Using reduced-size weight representations can help lower latency. However, this process introduces a quantization error, which can be significant. Various algorithms for quantization and de-quantization have been developed to minimize the impact of this error \cite{gaurav_efficiency_survey}.

Weight and activation quantization can be applied before or after training; for example, authors have taken large, pre-trained models and used a quantization policy to their weights and activations \cite{google_model_optimization}. This is an example of Post-Training Quantization. This leads to some amount of quantization error and can occasionally even increase inference time, especially if weights and activations are dynamically quantized and de-quantized (see Mobilenet-v2-1-224 in Table \ref{tab:quantization table} for an example of this effect). 

These advancements in quantization techniques are crucial for deploying LLMs on resource-constrained devices, reducing computational costs, and mitigating the environmental impact of large-scale AI systems \cite{langq2024}. Quantization can be applied before (QAT) or after training (PTQ). The strength of PTQ methods is that they do not require further data or training; a model can be trained, quantized, and deployed in a streamlined pipeline \cite{gaurav_efficiency_survey}. This is of the most significant benefit when considering large models that are impractical to re-train. On the other hand, QAT methods \cite{jacob2018quantization} generally result in better performance, as the model can adjust to the introduced quantization error but at the cost of additional training. This may be tractable for small models, but it is impractical for large transformer models with prohibitive training costs.

\subsection{Pruning}
\textcolor{black}{Pruning is a model compression technique that reduces the size of a neural network by removing unnecessary connections, aiming to improve computational efficiency while maintaining performance.}
It is well-understood that neural nets are over-parameterized to guarantee a path to a reasonable local loss minimum during training \cite{choromanska_loss_surfaces}. By dropping low-importance neurons or channels from the model, we can preserve high performance while minimizing the model's size and, thus, the resulting compute and energy costs.

\begin{figure*}[!ht]
    \centering
    \includegraphics[width=\linewidth]{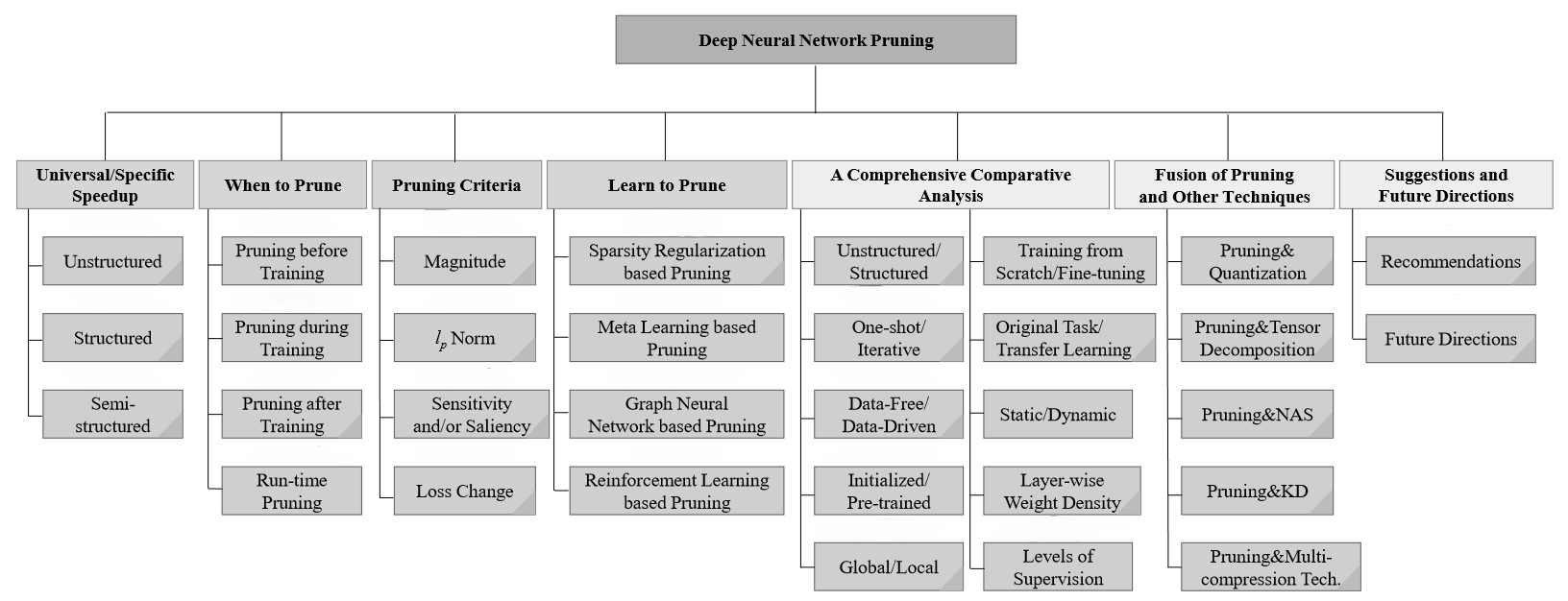}
    \vspace{-20pt}
    \caption{Pruning Taxonomy as formulated by \cite{pruning_survey_cheng}}
    \label{pruning_survey_hierarchy}
\end{figure*}

\textcolor{black}{Pruning techniques can be categorized by their timing (pre-training, during training, post-training), granularity (structured, unstructured), and decision criteria (rule-based, learning-based) as shown in Fig. \ref{pruning_survey_hierarchy}.}
Pruning methods can be classified based on their structural impact on deep neural networks (DNNs). Unstructured, or weight-wise pruning, is a fine-grained approach where individual weights in the network are pruned \cite{lee2019snipsingleshotnetworkpruning}. This method is efficient for small networks, as it reduces the overall number of weights by applying binary masks to the least essential parameters \cite{wang2020pickingwinningticketstraining}. In larger networks like LLMs, authors may merely set pruned weights to $0$ instead of maintaining a separate binary mask for all parameters \cite{frantar2023sparsegptmassivelanguagemodels}. Regardless, to realize the inference speed gains of a pruned network, specialized hardware or deployment-time compilers must be used to realize the gains from this fine-grained sparsification \cite{pruning_survey_cheng}. Structured pruning operates at a coarser level by removing groups of objects together: channels \cite{he2017channelpruningacceleratingdeep}, filters \cite{you_gate_2019}, transformer attention heads \cite{attention_head_pruning},  neurons \cite{ashkboos2024slicegptcompresslargelanguage}, layers \cite{men2024shortgptlayerslargelanguage},  and blocks \cite{ma2023llmprunerstructuralpruninglarge}. These structured methods offer inference acceleration regardless of hardware because the sparsification is coarse. If an entire channel is pruned, for example, then at inference, it is ignored; if instead the neurons in that channel are pruned in an unstructured way, the channel is still called at inference time, and the compiler and hardware must make sure to skip $0$ weight operations efficiently inside the retained channel. While structured pruning offers consistent inference acceleration across hardware, unstructured pruning often achieves higher compression ratios but requires specialized hardware or compiler optimizations to realize speedups. 

Structured pruning maximizes speed improvements by carefully selecting architectural elements to prune in a way that minimizes performance degradation.
Semi-structured pruning is a catch-all category for many methods that blend elements of structured and unstructured pruning \cite{xu2024lpvitlowpowersemistructuredpruning, ma2020imageenhancingpatternbasedsparsity, meng2020pruningfilterfilter}. By sequentially performing coarse pruning and then fine-grained pruning of the remaining structures, authors achieve greater compression levels while retaining performance.

As we mentioned, pruning seeks to decrease neural networks' size and computational expenses by eliminating weights or structural elements that minimally affect overall performance. Following this, we will introduce various pruning strategies—which differ mainly by the timing of their application—and emphasize significant research gaps.

\textbf{Pre-Training Pruning:} One forward-looking method is to prune networks based on randomly initialized weights \cite{wang2020pickingwinningticketstraining}. Though appealing for saving training time (because pruned weights require no subsequent computation), this approach risks issues like layer collapse \cite{tanaka2020pruningneuralnetworksdata}. 
It has been applied to convolutional neural nets (CNNs) \cite{lee2019snipsingleshotnetworkpruning,lee2020signalpropagationperspectivepruning}, but scaling it up to large models remains challenging, given the expense of training even pruned networks.

\textbf{Pruning During Training:} Another strategy embeds pruning into the training loop, iteratively updating which weights remain active. For instance, the RigL algorithm \cite{evci2021rigginglotterymakingtickets} periodically removes and regrows weights to maintain model capacity. Structured sparsity learning (SSL) \cite{wen2016learningstructuredsparsitydeep}, network slimming~\cite{liu2017_cnn_networkslimming}, and differentiable methods like Differential Sparsity Allocation (DSA) \cite{ning2020dsaefficientbudgetedpruning} similarly integrate pruning decisions with gradient-based updates. In neural architecture search, a related concept called Progressive Shrinking alternates between pruning and training to explore potential architectures~\cite{wang_apq_2020}. However, due to high computational costs, these techniques see limited exploration for large-scale models.

\textbf{Pruning After Training:} Pruning after training is widely used thanks to the abundance of large pre-trained models, which can be pruned and then fine-tuned on downstream tasks. Several algorithms focus on mitigating accuracy loss without retraining~\cite{frantar2023sparsegptmassivelanguagemodels,ashkboos2024slicegptcompresslargelanguage,kwon2022fastposttrainingpruningframework}, while others adopt a pipeline of pruning plus fine-tuning on a smaller dataset~\cite{liu2021groupfisherpruningpractical,ma2023llmprunerstructuralpruninglarge}. The Lottery Ticket Hypothesis~\cite{frankle2019lotterytickethypothesisfinding} is particularly influential in this space: it suggests dense, trained networks harbor sub-networks (``winning tickets'') that can match original performance at a fraction of the size. 
Research has confirmed these ideas in CNNs~\cite{you2022drawingearlybirdticketsefficient} 
and transformer-based language models~\cite{chen2021earlybertefficientberttraining}, 
including variations that prune large networks post-training (e.g., APQ~\cite{wang_apq_2020}) without full retraining. 
When retraining is needed but cost-prohibitive, lightweight fine-tuning schemes (e.g., LoRA) have shown promise for recovering performance~\cite{ma2023llmprunerstructuralpruninglarge}.

\textbf{Inference Time Pruning:} Inference time pruning (or dynamic pruning) adjusts the network \emph{per input}~\cite{rao2019runtimenetwork}, bypassing full computation for simpler samples. By pruning based on input complexity~\cite{tang2021manifoldregularizeddynamicnetwork}, resource usage can be reduced considerably without heavily compromising performance.

\textbf{Criteria for Pruning:} Different pruning criteria guide which weights or modules to remove. 
Magnitude-based methods prune weights below a certain threshold~\cite{han2016deepcompressioncompressingdeep}, 
while norm-based methods discard entire channels (e.g., using $L1$ norm~\cite{li2017pruningfiltersefficientconvnets,he2017channelpruningacceleratingdeep}). 
Alternative criteria include channel saliency~\cite{zhao2019variational}, neuron sensitivity~\cite{santacroce2023mattersstructuredpruninggenerative}, 
and module relevance~\cite{dery2024everybodyprunenowstructured}. Likewise, some work leverages structural graphs to locate highly connected components for more efficient pruning~\cite{ma2023llmprunerstructuralpruninglarge,zhang2021graphpruningmodelcompression}. 
Researchers have also introduced reinforcement learning to learn pruning policies~\cite{He_2018,rl_pruning} automatically.

Selecting a pruning strategy depends on model size, the target deployment scenario, and the available computational budget. 
Notably, while pruning yields smaller networks, it does not always guarantee proportional speedups on certain hardware backends. 
To address these gaps, a more quantitative outlook-comparing pruning to alternatives like quantization or tensor decomposition across metrics such as inference speed, memory footprint, and accuracy trade-offs-is needed. 
Such evaluations would be especially insightful for large-scale models that remain prohibitively expensive to retrain or fine-tune.

\subsection{Tensor Decomposition}
\begin{figure}[!ht]
    \centering
    \includegraphics[width=\linewidth]{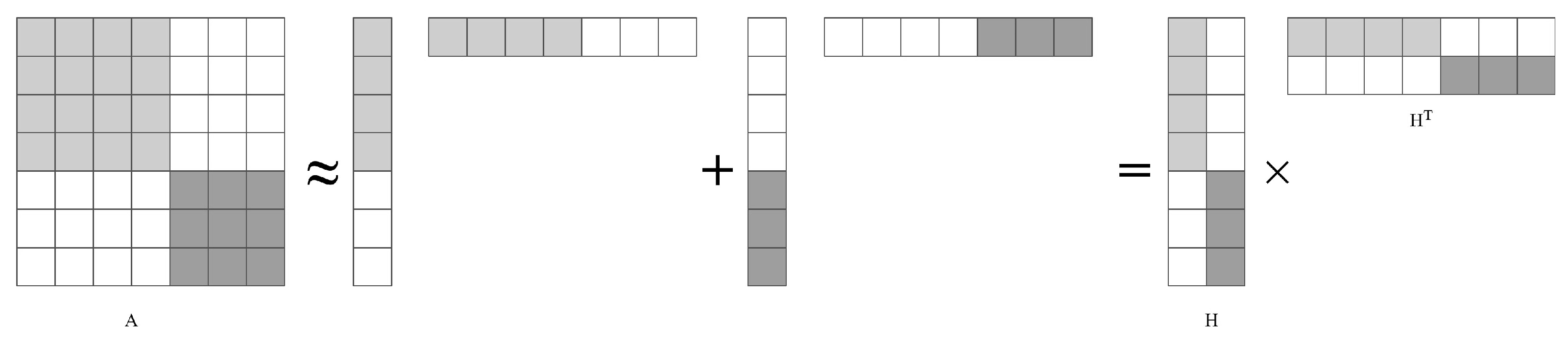}
    \vspace{-15pt}
    \caption{Kernel approximation of low-rank decomposition matrix as depicted in \cite{model_compression_survey}}
    \label{lrd_survey}
\end{figure}

Tensor decomposition is a recognized model compression method that approximates a high-rank weight tensor using products of lower-rank factors. This technique decreases the parameter count and computation costs associated with a neural network. As shown in Fig. \ref{lrd_survey}, substantial weight matrices and convolutional kernels can be broken down into smaller parts that effectively perform the same function, significantly reducing memory usage.

Historically, tensor decomposition was first applied to fully connected layers, factorizing large weight matrices into the product of two much smaller matrices. Subsequent research extended this idea to convolutional neural networks by decomposing large convolutional kernels into equivalent operations that increase the number of channels while reducing overall computation \cite{model_compression_survey}. Various decomposition strategies, such as CP, Tucker, and Tensor-Train Decomposition (TTD), provide different ways to balance representation capacity and resource efficiency.

In the context of foundation models, tensor decomposition shows particular promise. Because these models typically contain billions of parameters, decomposition can dramatically reduce memory usage and computational overhead, especially when paired with a short period of domain-specific fine-tuning \cite{low_rank_large_model, saha2024compressinglargelanguagemodels}. Recent research has focused on compressing large language models (LLMs) using tensor decomposition techniques. TensorGPT applies Tensor-Train Decomposition (TTD) to GPT-2 to compress token embeddings, achieving up to 65.64x compression ratios while maintaining comparable performance to the original model \cite{Mingxue2023}. MoDeGPT introduces a modular decomposition framework that partitions Transformer blocks into matrix pairs. It applies various matrix decomposition algorithms, achieving 25-30\% compression rates while maintaining 90-95\% zero-shot performance on LLAMA-2/3 and OPT models \cite{MoDeGPT2024}. These methods significantly reduce LLM size and computational requirements without substantial performance loss.

However, implementing decomposition-based compression on real-world hardware still poses challenges, including needing specialized kernels to maintain throughput gains. Future directions in tensor decomposition include automated rank selection, hybrid approaches that integrate decomposition with other compression methods (e.g., quantization and pruning), and rigorous exploration of decomposition within advanced architectures like Transformers.

\subsection{Knowledge Distillation}

\begin{figure}[!ht]
    \centering
    \includegraphics[width=250pt]{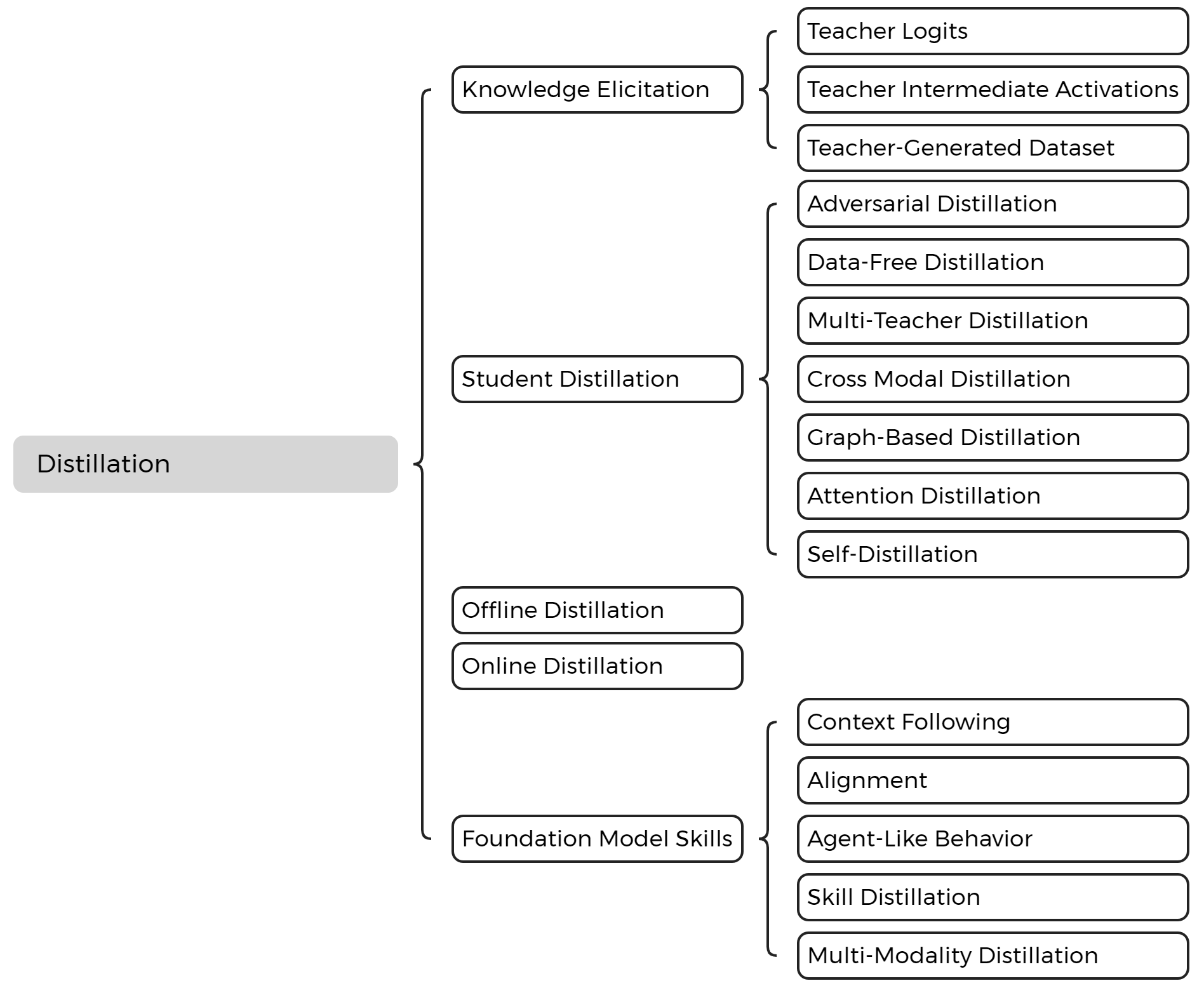}
    \vspace{-15pt}
    \caption{Distillation Components}
    \label{distillationmap}
\end{figure}

Distillation is a model compression technique that transfers knowledge from a large, complex model (teacher) to a smaller, efficient model (student) with minimal loss in performance, reducing resource costs and enabling deployment in resource-constrained environments \cite{hinton2015distillingknowledgeneuralnetwork}. Fig. \ref{distillationmap} referees to distillation components and approaches.


The two essential steps in knowledge distillation (KD) are knowledge elicitation and student distillation 
\cite{xu_survey_2024}. The initial step involves extracting knowledge from the teacher model. For instance, the teacher can be given input data to produce the corresponding output logits 
\cite{hinton2015distillingknowledgeneuralnetwork}. White-box methods, like feature extraction, provide direct access to the intermediate activations of the teacher model. In contrast, black-box techniques only utilize the output logits from both the teacher and student models 
\cite{xu_survey_2024}. Furthermore, in the context of foundation models, other knowledge elicitation strategies can be employed; the teacher might label a dataset for training the student, generate synthetic data through input expansion (creating new inputs based on initial seed data), or use data curation, where teacher feedback is applied to refine the dataset. After eliciting the knowledge, the student model is trained to mimic or approximate the teacher's knowledge. This training typically involves a loss function that minimizes the divergence between student and teacher logits but can also incorporate supervised fine-tuning (which maximizes the likelihood of sequences generated by the teacher) and reinforcement learning, where teacher feedback enhances the student's performance. These strategies aim to narrow the gap between the outputs of the student and teacher, facilitating effective knowledge transfer 
\cite{xu_survey_2024}.

Numerous knowledge distillation methods have been established, each with design considerations, techniques, and uses. 	\textbf{Adversarial Distillation} highlights that the compact model produced through distillation may be susceptible to adversarial threats \cite{Goldblum_adversarially_robust_distillation}
By employing the Generative Adversarial Network (GAN) framework, \cite{goodfellow2014generativeadversarialnetworks} researchers have created several training pathways that include Adversary, Teacher, and Student networks to enhance the final student’s robustness 	\cite{jung2024peeraidimprovingadversarialdistillation}. Additionally, GANs can be utilized for \textbf{Data-Free Distillation}, where the adversarial network produces synthetic examples, allowing the student to be trained without requiring extra data \cite{chen2019datafreelearningstudentnetworks}. Although adversarial distillation bolsters robustness, it frequently raises computational demands and training complexity, which may hinder scalability.


In \textbf{Multi-Teacher Distillation}, multiple teacher models distill their knowledge into a single student model. The knowledge is usually represented as logits or features, providing a comprehensive and balanced distillation from diverse sources \cite{hinton2015distillingknowledgeneuralnetwork, zhang2017deepmutuallearning}. Authors have developed alternative approaches by combining multi-teacher methods with Bayesian Neural Nets. For example, Vadera et al. show that a Bayesian Neural Net can generate an ensemble of teacher networks; this ensemble can then be used to train a student network as a multi-teacher distillation, with measurable benefits to the student's uncertainty calibration \cite{bayesian_distillation_jalaian}. \textbf{Cross-Modal Distillation} involves transferring knowledge from a teacher model trained on one modality to a student model designed to work with a different modality. Authors propose that students deployed on small or expensive-to-label datasets may benefit from this kind of transfer learning, as a large teacher model in one modality may incorporate the knowledge that improves the performance of the student in another modality \cite{gupta2015crossmodaldistillationsupervision}. Recent advancements include combining iterations of self-distillation with cross-modal data for regularization \cite{cross_modal_learning}, and filtering out cross-modal samples during distillation that do not add information to the student modality \cite{cross_modal_distillation}. In \textbf{Graph-Based Knowledge Distillation}, the data modality is restricted to graphs, and models to Graph Neural Networks (GNNs) \cite{liu2023graphbasedknowledgedistillationsurvey}. In systems that utilize attention mechanisms, like transformers, \cite{vaswani2023attentionneed} a distillation loss can be defined using the difference in attention mechanism between student and teacher; this is defined as \textbf{Attention-Based Distillation}. It has found use across modalities, including text \cite{self_attention_distillation} and images \cite{vision_attention_distillation}. The attention-based distillation loss is defined in Eq. \eqref{adl}.

\begin{equation}\label{adl}
    \mathcal{L}_{AT} = \frac{1}{A_h |x|} \sum_{a=1}^{A_h} \sum_{t=1}^{|x|} D_{KL} (\mathbf{A^T_{L,a,t}}|| \mathbf{A^S_{M,a,t}}),
\end{equation}

where $|x|$ and $A_h$ represent the sequence length and the number of attention heads; more details can be found in this paper \cite{self_attention_distillation}. The number of teacher and student network layers are denoted with $L$ and $M$; their last transformer layers are written $A^T_L$ and $A^S_M$.


This is not to be confused with other methods that define an activation-based distillation loss, also sometimes called attention \cite{crowley2019moonshinedistillingcheapconvolutions}, See Eq. \eqref{adl2}.

\begin{equation}\label{adl2}
    \mathcal{L}_{AT} =\mathcal{L}_{CE}(y,\sigma(s)) +\beta \sum_{i=1}^{N_L}\Big\|\frac{f(A_i^t)}{\|f(A_i^t)\|_2} - \frac{f(A_i^s)}{\|f(A_i^s)\|_2} \Big\|_2,
\end{equation}
where $s$ is the output logits of the student network. The cross-entropy loss, $\mathcal{L}_{CE}$, is added to the attention transfer component, which ensures that the difference between the spatial distributions of the student and teacher activations at selected layers in the network, $f(A^t_i)$, and $f(A^s_i)$, is minimized. This can be viewed as ensuring that the student network pays attention to the same things as the teacher network at those layers.

In addition to base model type and data modality, the distillation training task can also be defined in different ways. \textbf{Offline Distillation} is the conventional approach where the teacher model is pre-trained, and its knowledge is then used to guide the training of a smaller student model. \cite{hinton2015distillingknowledgeneuralnetwork} By contrast, \textbf{Online Distillation} occurs simultaneously, where both teacher and student models are trained together. This approach is appropriate when a high-capacity teacher model is unavailable, and knowledge needs to be distilled on the fly \cite{knowledge_distillation_survey}.
\textcolor{black}{This idea can even be extended to teacher-free frameworks, where ensembles of student models distill knowledge into each other simultaneously \cite{zhang2017deepmutuallearning, chen2019onlineknowledgedistillationdiverse}.
\textbf{Self-Distillation} involves the same model acting as both teacher and student, iteratively refining its own outputs. Some authors argue that self-distillation serves as a form of regularization, preventing overfitting by softening the output logits \cite{self_distillation_regularization}. Other researchers utilize self-distillation to reduce the model size by transferring knowledge from deeper layers to shallower layers through an iterative process \cite{self_distillation_zhang}.}

Given the size and scope of tasks for Foundation Models (FMs), there are numerous additional requirements defined by authors and various formulations of distillation to meet these requirements. In our survey, we adopt the framework proposed by Xu et al. \cite{knowledge_distillation_survey}, which encompasses a broad definition of distillation. For instance, if a teacher model labels a dataset used to train a student model, we consider this a form of distillation since the underlying knowledge is still transferred, even if specific distillation losses are not explicitly defined.

\textbf{Context Following} refers to the student model's ability to effectively interpret and respond to complex user inputs or contexts. Authors may view self-instruction as a form of self-distillation, where models generate their own input-output examples to further fine-tune themselves for specific tasks \cite{wang2022selfinstruct}. This approach can enhance the reasoning capabilities of smaller FMs by leveraging the strengths of larger models. Similarly, student models may select or filter data used by the teacher for fine-tuning, representing another form of self-distillation \cite{Li2024SelectiveRS}. Additionally, schemes like ORCA \cite{mukherjee2023orcaprogressivelearningcomplex} utilize step-by-step explanations from large foundation teacher models to train smaller models in reasoning through tasks. For conversational models, self-distillation with feedback involves the large teacher model ranking the utility of multiple proposed outputs from the small model, using this feedback to refine the small model's responses \cite{xu2023baizeopensourcechatmodel}. Recently, Knowledge-Augmented Reasoning Distillation (KARD) \cite{kang2023knowledgeaugmentedreasoningdistillationsmall} has been introduced to address the challenge of smaller FMs memorizing vast amounts of data required for expert-level question answering. KARD involves retrieving data from external sources and using it to fine-tune the small model's responses, enhancing both distillation for small FMs and the performance of general-purpose models \cite{asai2023selfraglearningretrievegenerate}.

\textbf{Alignment} pertains to the qualitative ability of a machine learning algorithm to understand and respond in accordance with human-intuitive values. Recent benchmarks for this quality in FMs include MT-Bench \cite{mt_bench} and AlpacaEval \cite{alpaca_eval}. In this context, distillation into a smaller model has been observed to decrease alignment with user intent \cite{tunstall2023zephyrdirectdistillationlm}. Ongoing efforts aim to develop methods for distilling teacher-model alignment into student models and enhancing the alignment of the resulting students. For example, Anthropic's Constitutional AI \cite{bai2022constitutionalaiharmlessnessai} evaluates responses based on a concise set of human-generated rules to ensure alignment.

\textbf{Agent-Like Behavior} involves enhancing the autonomous capabilities of the student model, enabling it to plan and execute actions in response to inputs by utilizing external``tools" (e.g., other models or APIs) to assist in task completion. One approach involves distilling two student models: the first, a `sub-goal generator', takes the last ten actions of the agent and an overarching task as input to output the current sub-goal. The second student, an `action generator', takes the current sub-goal, overarching task, and last ten actions to predict the next action \cite{sub_goal_distillation}. Similar methodologies are found in the Lumos framework \cite{yin2024agentlumosunifiedmodular} and the FireAct framework \cite{chen2023fireactlanguageagentfinetuning}, which utilize a limited number of agent-action trajectories for fine-tuning. Other researchers have focused on enabling FMs to accurately call APIs, minimize hallucinations, and remain robust to distribution shifts over time in API calls by incorporating distillation techniques specific to API-call tasks \cite{patil2023gorillalargelanguagemodel}.

\textbf{Skill Distillation} enables the student model to specialize in a particular domain, such as Natural Language Processing (NLP) or computer vision. Training a general-purpose student model using a large teacher across extensive datasets is often infeasible due to time and cost constraints. However, by carefully selecting a smaller, domain-specific data distribution, relevant skills can be effectively transferred to the student. Efforts to streamline the distillation process include MiniLLM \cite{gu2024minillmknowledgedistillationlarge} and DistillLLM \cite{ko2024distillmstreamlineddistillationlarge}. Alternative methods, such as Impossible Distillation \cite{jung2023impossible}, use paraphrastic proximity to retrieve high-quality paraphrase datasets and distill student models that outperform their GPT-2 teachers, providing both datasets and trained student models. In NLP search and data augmentation tasks, QUILL employs a two-stage distillation process where a `Professor' model generates long, retrieval-augmented responses. These responses are distilled into a `Teacher' model without retrieval capabilities, which then annotates an unlabeled dataset for distillation into the final 'Student' model \cite{srinivasan2022quillqueryintentlarge}. Skill distillation can also be performed across different data modalities, similar to non-FM approaches.

\textbf{Multi-Modality Distillation} involves transferring knowledge from a teacher model operating in one modality to a student model in a different modality (e.g., visual to textual). This is typically achieved through the teacher model's labeling of a dataset, which the student model then uses for training. Notable examples of such methods include LLaVa \cite{liu2023visualinstructiontuning} and Macaw-LLM \cite{lyu2023macawllmmultimodallanguagemodeling}.

Model compression addresses researchers' needs for streamlined search processes and rapid deployment. However, there are scenarios where achieving higher performance necessitates more advanced architectures beyond compression. In such cases, it becomes essential to employ methods that can search within an appropriate architectural space to optimize specified objectives while adhering to imposed constraints, which are often dictated by hardware limitations and mission requirements at the edge.
This necessity leads to the exploration of \textbf{Neural Architecture Search} (NAS), a pivotal area in model optimization that systematically designs and identifies optimal neural network architectures tailored to specific tasks and constraints.

\textbf{Pruning vs.\ Quantization:} Recent studies argue that quantization may, in some cases, be superior to pruning~\cite{yin2024junk}. Weights deemed unimportant in one domain might still be crucial for challenging downstream tasks, 
and quantizing them (rather than removing them) can better preserve performance under resource constraints. 

\section{Neural Architecture Search}
\label{sec:nas}

\begin{figure}[ht]
    \centering
    \includegraphics[scale=0.25]{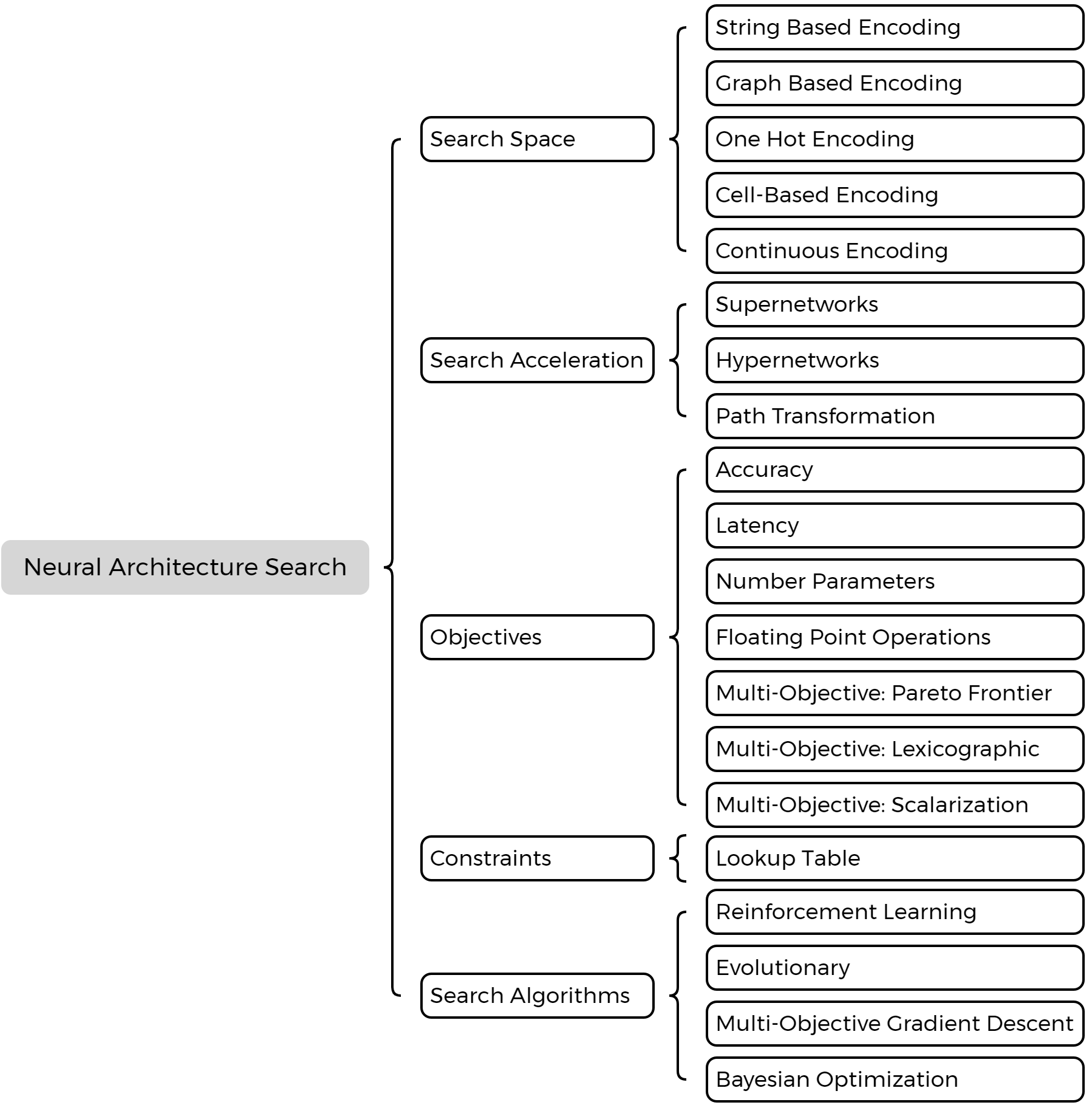}
    \caption{Concepts and Tools in Neural Architecture Search}
    \label{NASmap}
\end{figure}

Neural Architecture Search (NAS) automates the design of neural network architectures by systematically exploring a search space of possible configurations. As highlighted by \cite{he_automl_2021}, NAS is a sub-field of Automated Machine Learning (AutoML) specializing in neural architectures. Fig. \ref{NASmap} outlines concepts and tools related to NAS. It involves:
\begin{enumerate}
    \item \textbf{Defining the search space} (e.g., convolutional, recurrent, or fully connected layers),
    \item \textbf{Selecting a search strategy} (or Architecture Optimization method),
    \item \textbf{Choosing an evaluation method} to gauge candidate architectures' performance.
\end{enumerate}
Additionally, researchers often distinguish \textbf{architecture optimization (AO)} from \textbf{hyperparameter optimization (HPO)}: AO concerns layer configurations and connectivity, while HPO adjusts non-architectural factors such as batch size or learning rate.

\subsection{Defining the Search Space}

\textbf{Neural Architecture Search} begins with specifying which architectures are valid candidates. For instance, the architecture might be a sequence of convolutional and pooling layers, or a more complex graph of connected modules. This definition greatly impacts both the quality of results and the computational cost of NAS.

\paragraph{Primitive vs.~Composite Components}
Researchers can allow only a set of primitive operations (e.g., ``3$\times$3 convolution, ReLU, max pooling''), creating a large, expressive search space \cite{zoph_neural_2017,cai_efficient_2017}. Such expressiveness can discover novel architectures but may demand intensive computation. Conversely, a higher-level ``cell-based'' or ``motif-based'' approach restricts the search space to composite building blocks (cells), greatly reducing complexity but potentially missing innovative designs.

\begin{figure*}[ht]
    \centering
    \includegraphics[scale=0.6]{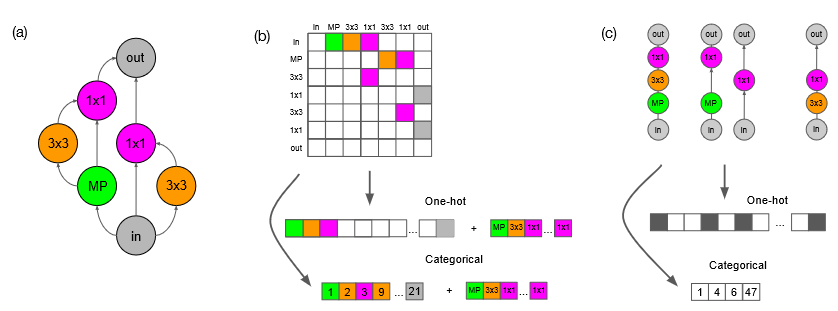}
    \caption{A Study on Encodings for NAS \cite{white_study_encodings}. (a) A CNN visualized as a DAG; (b) Adjacency matrix of the DAG with one-hot and categorical encodings; (c) A path-based graph representation.}
    \label{nas_encoding}
\end{figure*}

\paragraph{Leveraging Encodings}
An effective way to limit or navigate the search space is by encoding architectures into concise representations:

\begin{itemize}
    \item \textbf{String-Based Encoding}: The network is represented as a sequence of tokens denoting operations and hyperparameters (e.g., \texttt{[Conv 32 3x3, ReLU, MaxPool, Dense 128]}) \cite{zoph_neural_2017,cai_efficient_2017}.
    \item \textbf{Graph-Based Encoding}: A directed acyclic graph (DAG) describes data flow, with edges as operations and nodes as data tensors (or vice versa). This DAG can be flattened as an adjacency matrix or represented path-by-path (see Fig.~\ref{nas_encoding}) \cite{white_study_encodings}.
    \item \textbf{Cell-Based Encoding}: Instead of searching large architectures at once, one can \emph{first} search for a relatively small ``cell'' (a DAG of operations) and then stack or concatenate multiple copies of this cell to form the full network \cite{white2023neuralarchitecturesearchinsights}. This hierarchical approach is especially popular for large models, reducing a potentially massive search space to a smaller cell space plus macroscale configuration.
\end{itemize}

\paragraph{Differentiable Architectures}
While many search spaces are discrete, \textbf{differentiable NAS} (e.g., DARTS \cite{darts_liu}) continuously parameterizes architectural choices. For a DAG with multiple candidate operations at each edge, differentiable weights determine which operations are active. Training these weights via gradient descent dramatically accelerates the search but can produce suboptimal or unstable results if the relaxed approximation is biased.

\paragraph{Abstract Architectures}
Recent work notes that many architectures with similar global properties (e.g., total depth or width) often exhibit similar performance \cite{program_synthesis_nas,approximate_properties}. Hence, some authors reduce the search space to high-level abstractions, evaluating only one representative network per abstract configuration.

\subsection{Search Strategies}

\paragraph{Supernetworks}
\textbf{Once-For-All (OFA) training} \cite{cai_once-for-all_2020} is a widely-known supernetwork approach. One trains a large model that conceptually contains multiple sub-networks (created by pruning layers or channels). Each sub-network is then evaluated with minimal additional fine-tuning, drastically cutting the repeated training cost. This idea is related to \textbf{Progressive Shrinking}, where the model is iteratively pruned and fine-tuned to preserve performance \cite{wang_apq_2020}. However, supernetworks can be impractical for extremely large architectures (e.g., LLMs) because the combined supernetwork might exceed feasible memory or training time \cite{acc_predictor_limitations}. Moreover, if the supernetwork's weight-sharing scheme is biased, the sub-network performance can be noisy or misleading.

\begin{figure*}[ht]
    \centering
    \includegraphics[scale=0.45]{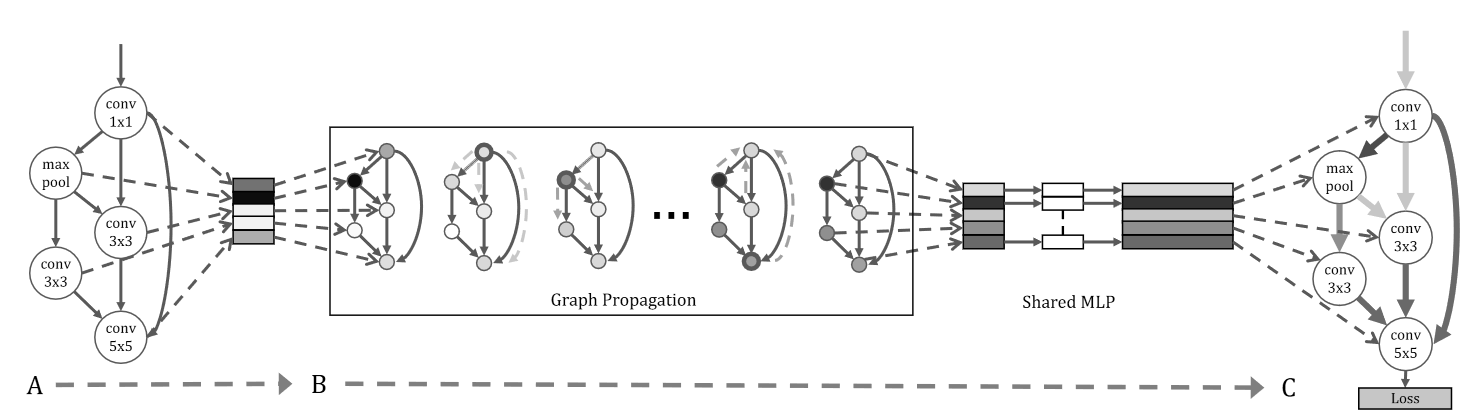}
    \vspace{-5pt}
    \caption{Graph Hypernetwork framework \cite{graph_hypernetworks}. 
    \textbf{(A)} A neural network architecture is sampled and input to a GHN.
    \textbf{(B)} The GHN performs graph propagation to generate weights.
    \textbf{(C)} The GHN is trained to minimize the training loss of the generated network, ranking architectures by performance.}
    \label{fig:graph_hypernetwork}
\end{figure*}

\paragraph{Hypernetworks}
A \textbf{hypernetwork} predicts the weights for candidate architectures, bypassing the need to train each from scratch \cite{li2020dhpdifferentiablemetapruning, liu2019metapruningmetalearningautomatic}. 
\emph{Graph Hypernetworks (GHNs)} \cite{graph_hypernetworks} (see Fig. \ref{fig:graph_hypernetwork}) extend this to a graph neural network (GNN) that reads an input DAG (the candidate architecture) and outputs all free parameters for that network. Evaluating many architectures thus involves:
\begin{enumerate}
    \item Feeding each architecture’s graph into the GHN;
    \item Generating the corresponding weights;
    \item Quickly measuring its performance (e.g., on a small validation set).
\end{enumerate}
Since the GHN itself is trained only once, large-scale searches become more computationally tractable.

\paragraph{Path-Transformation}
In \textbf{path-transformation} \cite{cai_path-level_2018}, one starts with a large model and gradually transforms it by adding or pruning layers, guided by reinforcement learning (RL) or heuristic rules. Weight sharing between unchanged components avoids retraining from scratch each time. After each transformation, the model is quickly fine-tuned; the search strategy monitors validation metrics to decide the next step.

\subsection{Objectives in NAS}\label{subsec:objectives}

NAS typically seeks to maximize performance (\emph{e.g.}, accuracy) and/or minimize resource use (\emph{e.g.}, latency, memory). However, fully training \emph{each} candidate to measure accuracy is often impractical with a large search space. Hence, authors have proposed multiple \emph{surrogates} or \emph{predictors} to approximate accuracy.

\paragraph{Accuracy Predictors and Early Stopping}
Instead of full training, an \textbf{accuracy predictor} \cite{wang_apq_2020,cai_once-for-all_2020} can take an architecture encoding as input and output an estimated accuracy. The predictor itself is learned from a small subset of architectures that have been fully trained. Alternatively, \textbf{early stopping} techniques\cite{auto-distill} use partial training to guess final accuracy, which cuts search time. However, these predictions can be highly variable if the underlying loss surface is rugged or if the training set for predictor modeling is too small \cite{choromanska_loss_surfaces,acc_predictor_limitations,hutter_performance_predictors}.

\paragraph{Latency and Parameters}
\textbf{Latency} (inference time) is crucial for real-time tasks. Yet, parameter count (or FLOPs) and latency do not necessarily correlate on modern hardware \cite{flop_acceleration_li,auto-distill}. In some cases, a model with more parameters can run faster if its architecture aligns better with a device’s parallelization. Consequently, many researchers measure latency on target hardware, use a \textbf{lookup table}, or train a \textbf{latency predictor} model for quick estimates \cite{MNASNET_tan,wang_apq_2020}.

\paragraph{Energy and Robustness}
In battery-limited edge environments, \textbf{energy consumption} can be a direct objective \cite{tensorrt_energy}. Other specialized objectives include \textbf{adversarial robustness} \cite{wu2024robustnasadversarialtraining} and \textbf{uncertainty calibration}, depending on domain requirements.

\paragraph{Neural Scaling Laws}
Large-scale studies \cite{kaplan2020scaling,bahri2024explaining} show that performance often follows empirical \emph{scaling laws} w.r.t.\ data size ($D$), compute ($C$), and parameter count ($N$). While these laws do not prescribe a specific search method, they reveal diminishing returns beyond certain scales.

\subsection{Multi-Objective Optimization and the Pareto Frontier}

Real-world applications frequently require \textbf{multi-objective optimization (MOO)}. For instance, accuracy, latency, and memory can all matter simultaneously. A configuration is \emph{Pareto-optimal} if no objective can be improved without worsening another \cite{freitas_opinion_moo_hybrid}.

\paragraph{Pareto Frontier}
Collectively, all Pareto-optimal solutions form the \textbf{Pareto frontier}. Practitioners often select from these trade-off points according to specific hardware or mission needs.

\paragraph{Lexicographic Optimization}
Some approaches prioritize objectives in a strict hierarchy \cite{lexicographic_opt}. For example, one might first maximize accuracy, then among top candidates minimize latency, and only then consider FLOPs. This method yields a single final solution guaranteed to lie on the Pareto frontier but might ignore equally valid trade-offs.

\paragraph{Scalarization}
Alternatively, multiple objectives can be merged into a single metric via additive or multiplicative factors. MnasNet \cite{MNASNET_tan}, for instance, uses a function that blends accuracy with latency constraints. One drawback is that scalarization can bias solutions toward certain regions of the Pareto frontier \cite{auto-distill}.

\paragraph{Hybrid Approaches}
Some authors combine lexicographic methods with scalarization or even dynamically shift objective weights during search \cite{freitas_opinion_moo_hybrid}. This can systematically explore multiple zones of the frontier.

\subsection{Constraints}

Apart from explicit objectives, constraints (e.g., hardware memory limits, real-time latency caps) often further restrict NAS. A common technique is the \textbf{lookup table (LUT)} approach \cite{wang_apq_2020}, which stores per-layer or per-block cost (latency, energy, etc.) on the target hardware. The total cost of a candidate architecture is approximated by summing the LUT entries. While computationally cheap, LUT-based sums can be imprecise when concurrency or caching effects become significant \cite{MNASNET_tan}. Hence, LUT methods typically function as a quick screening step to discard obviously infeasible designs.

\subsection{NAS Optimization Algorithms}

Given the (often large) search space and potential multi-objective setting, authors have proposed various optimization algorithms:

\paragraph{Reinforcement Learning (RL)}
Early NAS work by Zoph and Le \cite{zoph_neural_2017} used a policy-gradient RL approach to iteratively propose architectures, receiving rewards based on validation accuracy. This framework naturally extends to \textbf{pruning} or \textbf{quantization} decisions \cite{rl_pruning,He_2018} by defining actions (e.g., ``remove 10\% of weights'') and rewards (e.g., trade-off in accuracy vs.\ model size). Parameter sharing \cite{pham2018_rl_paramatersharing} can reduce redundant computation among candidates.

\paragraph{Evolutionary Algorithms}
Evolutionary strategies view each architecture as an organism in a population \cite{evolutionary_nas,large_evolutionary_nas}. After evaluating each architecture’s metrics, top performers ``survive'' to the next generation, while new ones are created by mutation (altering some layers/operations) or crossover (combining parts of two architectures).

\paragraph{Multi-Objective Gradient Descent}
When architectures can be continuously relaxed (as in DARTS \cite{darts_liu}), \textbf{multi-objective gradient descent} \cite{desideri_multiple-gradient_2012,hutter_moo_differentiable_nas} can simultaneously optimize multiple objectives. This requires carefully balancing gradients for each objective at each iteration. However, pure differentiable methods may struggle with the inherently discrete nature of many architectural choices.

\paragraph{Bayesian Optimization (BO)}
In the context of NAS + HPO, a \textbf{surrogate model} predicts performance from an architecture encoding; an \emph{acquisition function} (e.g., expected improvement, upper confidence bound) balances exploration of uncertain regions against exploitation of promising areas. BANANAS \cite{white_bananas}, AutoDistill \cite{auto-distill}, and Transfer NAS \cite{transfer_bayesian_nas} exemplify BO for NAS. Notably, BO can maintain uncertainty estimates over the performance predictor to adaptively guide the search toward regions most likely to yield better Pareto-optimal architectures.

\vspace{0.5em}
\noindent
In summary, \textbf{Neural Architecture Search} offers a systematic way to discover efficient models under diverse objectives (accuracy, latency, energy) and constraints (memory, real-time performance). By carefully defining the search space, leveraging surrogates or hypernetworks to reduce training costs, and choosing suitable algorithms, NAS can automate the design of neural networks for resource-constrained edge computing scenarios.

\section{Compiler and Deployment Frameworks}\label{sec:compiler_deployment}

While model compression and neural architecture search aim to yield lightweight networks, additional \emph{deployment-specific} optimizations can be achieved via specialized compilers and hardware platforms. These compilers transform trained models into optimized code that exploits hardware-specific features (e.g., tensor cores, vector units, sparse processing) for faster inference and lower power consumption. Below, we summarize several notable solutions.

\subsection{Industry-Developed Compilers}

\paragraph{TVM}
\textbf{Tensor Virtual Machine (TVM)} \cite{chen2018tvm} is an open-source compiler that parses a trained neural network model into a hardware-agnostic computational graph. It then applies graph-level optimizations (e.g., operator fusion) and generates low-level code tuned to the target hardware. Experiments show speedups of 1.2--3.8$\times$ on CPU, GPU, and FPGA backends.

\paragraph{XLA}
\textbf{Accelerated Linear Algebra (XLA)}\footnote{\url{https://github.com/openxla/xla}} is another open-source compiler (primarily developed by Google) that similarly decomposes a model into a high-level graph. It fuses operations into hardware-aware kernels, often achieving lower latency than unoptimized TensorFlow or JAX execution.

\paragraph{GroqFlow}
\textbf{Groq} provides custom hardware based on the Tensor-Streaming Processor (TSP) architecture \cite{gwennap_groq_2020}. Its compiler, GroqFlow, converts trained models into programs for Groq devices. Internal benchmarks report a 2$\times$ increase in images-per-second on ResNet-50 compared to an Nvidia V100 GPU.

\paragraph{OpenVINO}
\textbf{OpenVINO} \cite{intel_openvino} is Intel’s open-source toolkit that includes hardware-aware optimizations for CPUs, GPUs, and other Intel accelerators. It also ships with a \emph{Neural Network Compression Framework (NNCF)} supporting pruning and quantization (both post-training and during training). This integration can further reduce inference time and memory footprint on Intel platforms.

\paragraph{TensorRT}
\textbf{TensorRT} \cite{tensorrt} is Nvidia’s open-source compiler for its GPUs and specialized accelerators. It offers optimizations for quantization and structured sparsity, and includes \textbf{TensorRT-LLM} for large language models. Empirical results suggest at least a 50\% reduction in latency and energy usage for LLM inference \cite{tensorrt_energy}.

\subsection{Specialized Hardware for Edge AI: IBM NorthPole}

Beyond general-purpose compilers, certain vendors design hardware specifically for \emph{edge} inference. \textbf{IBM NorthPole} \cite{ibm_northpole} is a neuromorphic chip that uses event-driven processing to achieve high energy efficiency. By activating computation only on relevant data and pruning unnecessary paths, it significantly reduces power draw. On-chip memory further cuts down latency and off-chip transfers. NorthPole also supports weight pruning and quantization at the hardware level, compressing model layers and thereby accelerating inference. This combination of sparse data processing, event-driven operation, and local memory makes NorthPole particularly suitable for low-power, real-time tasks in edge environments.

\subsection{Relation to the Overall Pipeline}

These compilers and specialized hardware are often the final step in a deployment pipeline. Even after \emph{model compression} (Section~\ref{sec:model-compression}) or \emph{NAS} (Section~\ref{sec:nas}), the resulting model can benefit further from hardware-tailored graph optimizations. Consequently, the choice of compiler or hardware backend directly impacts whether or not the theoretical speedups from compression or NAS fully manifest in real-world inference. Practitioners aiming for efficient edge deployment typically combine all three: 
\begin{enumerate}
    \item Model compression or NAS to produce a compact, high-performing architecture,
    \item Possibly re-quantizing or pruning for the specific device,
    \item Compiling with a toolchain (TVM, TensorRT, etc.) specialized for the target hardware.
\end{enumerate}

In summary, compiler frameworks and specialized hardware backends constitute a critical link in the optimization chain, translating theoretical gains from compression and architecture search into tangible, real-time performance improvements.

\section{Integrated Approaches and Case Studies}\label{sec:integrated_approaches}

In this section, we first discuss common \emph{experimental methods} and publicly available \emph{NAS benchmarks} that facilitate reproducible research. We then highlight several \emph{case studies} and \emph{integrated approaches} that combine the design elements introduced earlier: compression (pruning, quantization, distillation), NAS + HPO, and multi-objective optimization. The examples illustrate how these techniques work together to meet the unique demands of edge computing.

\subsection{Common NAS Benchmarks}

\textbf{NAS Benchmarks} provide standardized environments for evaluating and comparing different neural architecture search algorithms. Such benchmarks pair a fixed \emph{search space} with tasks and often include pre-computed results or surrogates to minimize the computational overhead of repeatedly training candidate architectures.

\paragraph{NAS-Bench-101}
Proposed in \cite{nasbench_101}, \textbf{NAS-Bench-101} targets the CIFAR-10 dataset, representing each candidate architecture as a Directed Acyclic Graph (DAG) with up to 7 vertices and 9 edges. Allowed operations are $3\times 3$ convolution, $1\times 1$ convolution, and $3\times 3$ max pooling. Each valid DAG is then repeated (stacked) three times to form the final network. This search space encompasses $\sim 423,\!000$ distinct architectures after excluding duplicates or invalid topologies. For each architecture, the authors trained 3 random initializations to convergence and averaged the final accuracy. This yields a \emph{lookup table} mapping \{architecture $\to$ accuracy, training time\}, enabling researchers to compare NAS algorithms without the heavy cost of full model training each time. However, NAS-Bench-101 only provides \emph{single-objective} data (accuracy and runtime), limiting its direct utility for multi-objective (e.g.\ accuracy-latency) research.

\begin{figure}[!ht]
    \centering
    \includegraphics[width=0.5\linewidth]{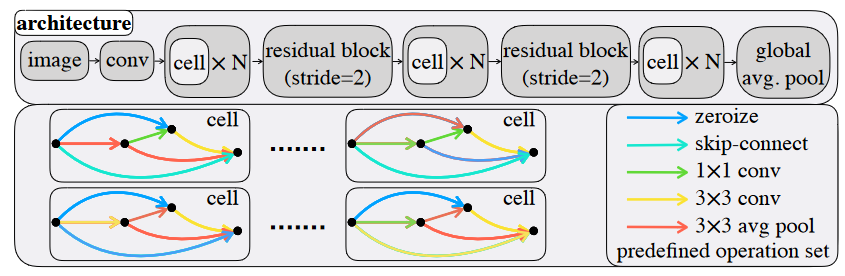}
    \caption{NAS-Bench-201: Each candidate cell has four nodes, and the overall network consists of five stacked cells \cite{nasbench_201}.}
    \label{NASBench201}
\end{figure}

\paragraph{NAS-Bench-201}
\cite{nasbench_201} extend the idea to three datasets: CIFAR-10, CIFAR-100, and ImageNet-16-20 (see Fig. \ref{NASBench201}). This benchmark defines a more compact search space of about 15,600 architectures. Each is formed by stacking five identical cells, where each cell has four internal nodes and five possible operations. Performance metrics include model accuracy, latency, floating-point operations (FLOPs), parameter count, and training time. Notably, NAS-Bench-201 also provides full learning curves over 200 epochs, facilitating deeper analyses of training dynamics. Its main limitation is that the architectures themselves are still relatively small compared to modern large-scale applications.

\paragraph{NAS-Bench-301}
Recognizing that NAS-Bench-101 and 201 cover relatively small design spaces, \textbf{NAS-Bench-301} \cite{nasbench_301} takes a different approach. Instead of enumerating all possible architectures and storing their trained results in a table, it trains a \emph{surrogate model} to predict performance in the \emph{DARTS-based} search space on CIFAR-10. This surrogate leverages an ensemble of Graph Isomorphism Networks (GINs) to predict accuracy and an additional LGBoost-based predictor for latency. The dataset includes $\sim 60,\!000$ fully trained architectures. By sampling new designs and querying the surrogate, researchers can approximate performance quickly, enabling multi-objective or hardware-aware NAS studies at larger scales than NAS-Bench-101/201.

\paragraph{NAS-Bench-Suite}
Finally, \cite{nasbench_suite_hutter} aggregate multiple NAS benchmarks (including the ones above) into a single open-source repository (\texttt{NasLib}\footnote{\url{https://github.com/automl/naslib}}). The authors observe that approaches or hyperparameters that excel in one benchmark may not generalize to others. Consequently, NAS-Bench-Suite helps evaluate NAS algorithms more robustly across different tasks and search spaces.

\subsection{Case Studies Integrating Compression and NAS}

With the foundational benchmarks in mind, we now examine several integrated approaches. These methods combine elements of \textbf{model compression} (pruning, quantization, distillation), \textbf{NAS + HPO}, and \textbf{hardware constraints} to achieve efficient inference on resource-limited devices.

\subsubsection{APQ: Pruning, Quantization, and Supernetworks \cite{wang_apq_2020}}
\label{subsec:apq}

\begin{figure}[!ht]
    \centering
    \includegraphics[width=0.7\linewidth]{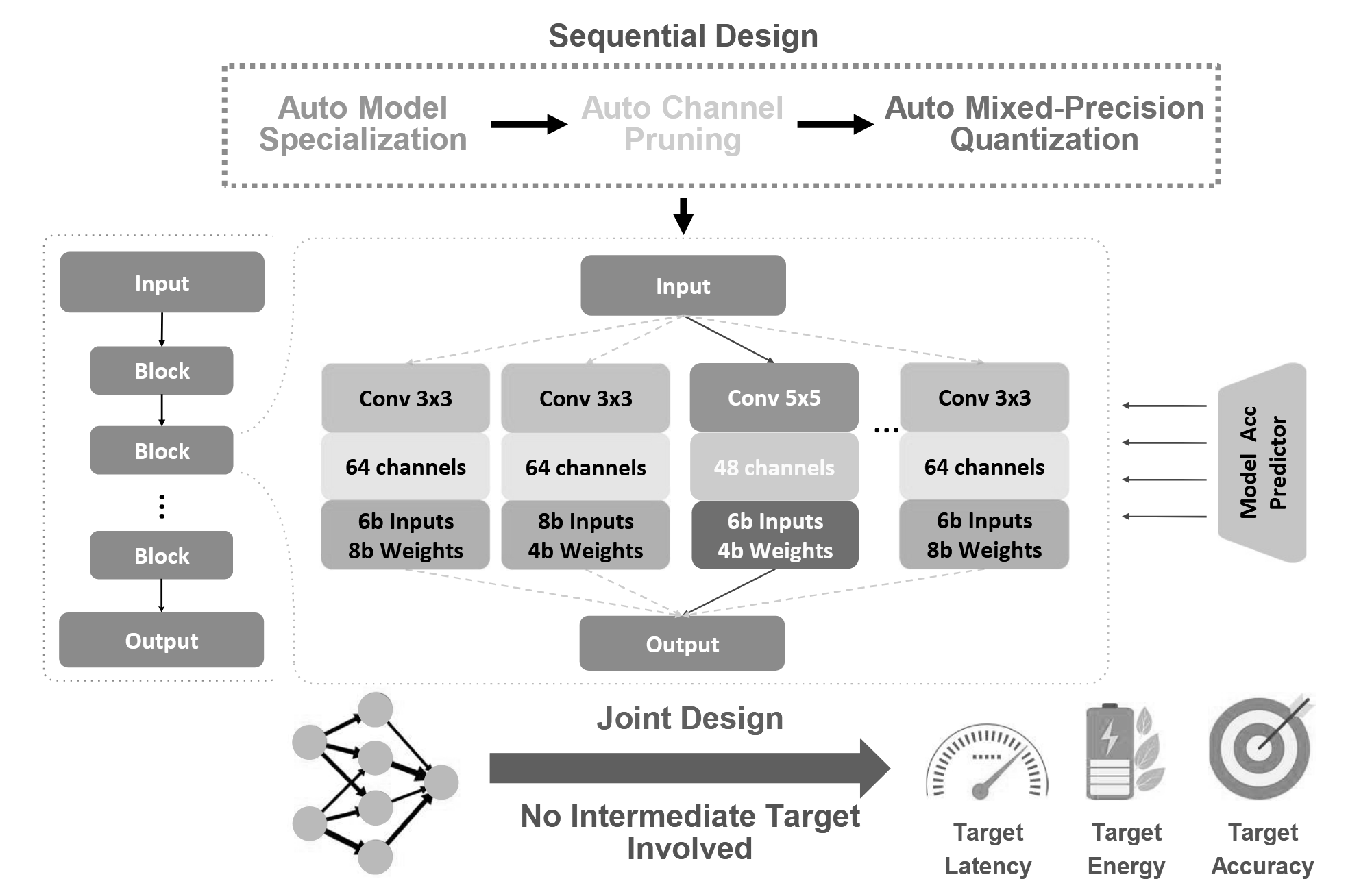}
    \vspace{-10pt}
    \caption{APQ Framework \cite{wang_apq_2020}. The supernetwork (left) supports multiple sub-architectures via pruning, and each sub-architecture can have different quantization policies. An accuracy predictor plus hardware LUT guide an evolutionary search.}
    \label{APQ Framework}
\end{figure}

\textbf{APQ} (\emph{Accuracy Predictor for Quantization}) unifies pruning, quantization, supernetwork-based NAS, multi-objective optimization, and a hardware-driven \emph{lookup table} approach (see Fig. \ref{APQ Framework}):
\begin{enumerate}
    \item \textbf{Supernetwork Training}: First, a large convolutional neural network (CNN) supernetwork is trained on an image classification task. This supernetwork implicitly contains many sub-architectures (via pruning).
    \item \textbf{Progressive Shrinking and Validation}: By pruning different subsets of channels or layers, one obtains sub-architectures, each with its own validation accuracy. Encoding each sub-architecture as a bit-vector plus its measured accuracy builds a dataset of \{architecture, accuracy\} pairs.
    \item \textbf{Quantization Policies}: The authors apply different quantization policies (e.g., 8-bit, mixed precision) to those sub-architectures and record resulting accuracies. These tuples train an \emph{accuracy predictor} that simultaneously captures the impact of pruning \emph{and} quantization.
    \item \textbf{Evolutionary Search}: With this accuracy predictor, APQ searches over the joint space of architecture (pruning) and quantization policies. 
    \item \textbf{Lookup Table Constraints}: Finally, a hardware LUT (layer-level latency and energy estimates) is summed for each candidate. Subnetworks exceeding allowable constraints are discarded. This yields a Pareto set of feasible (architecture, quantization) solutions.
\end{enumerate}

Experiments show that APQ can effectively find high-accuracy, low-latency CNNs by combining these modular compression and search steps in a single pipeline.

\subsubsection{DARTS: Differentiable Architecture Search \cite{darts_liu}}
\label{subsec:darts}

\textbf{Differentiable ARchiTecture Search (DARTS)} tackles NAS by relaxing discrete architecture choices into continuous, learnable parameters. Concretely:
\begin{itemize}
    \item A DAG is defined where each edge can be one of several operations (e.g., $3\times3$ conv, $5\times5$ conv, skip connection). 
    \item Real-valued coefficients (architecture parameters) weigh these operations.
    \item A \emph{bilevel} optimization scheme alternates updating \emph{model weights} (on the training set) and \emph{architecture parameters} (on the validation set).
\end{itemize}

\begin{figure}[!ht]
    \centering
    \includegraphics[width=0.7\linewidth]{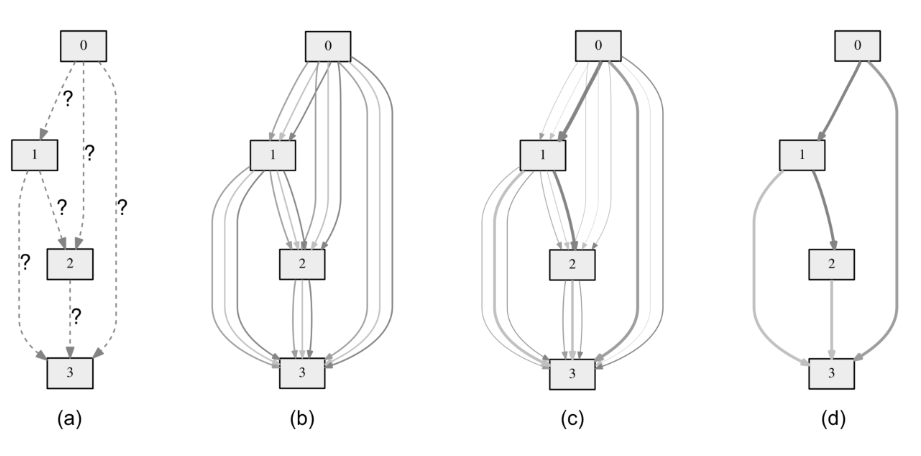}
    \vspace{-15pt}
    \caption{DARTS Framework \cite{darts_liu}. A DAG with multiple candidate operations per edge is learned via continuous weights, then discretized after training.}
    \label{darts_framework}
\end{figure}

After convergence, the operation with the highest coefficient on each edge is selected, yielding a discrete architecture (See Fig. \ref{darts_framework}). The authors demonstrate that DARTS can reduce search time for CIFAR-10 from thousands of GPU-days (NASNet-A) to about 4 GPU-days, drastically lowering the computational barrier.

Notably, DARTS itself does not inherently integrate pruning or quantization, but variants have extended the differentiable approach to incorporate these techniques.

\subsubsection{AWQ: Activation-Aware Weight Quantization \cite{lin_awq_2024}}
\label{subsec:awq}

\textbf{AWQ} focuses on compressing large transformer-based models via \emph{channel-level quantization}. Observing that most weights can be quantized aggressively with minimal loss, the authors design a scheme to treat only the top $\sim1\%$ of high-activation channels with higher precision:
\begin{enumerate}
    \item Measure \emph{average activation} per channel on a validation set.
    \item Retain full precision for the channels with the largest activations (e.g., top 1\%), quantize the rest to 4-bit floats.
    \item Optionally allow fine-tuning if training data is available.
\end{enumerate}
This selective strategy yields speedups of 3--4$\times$ for large LLMs (e.g., LLaMA, OPT) while preserving perplexity or accuracy on benchmarks (captioning tasks, visual-language tasks, etc.). Latency improvements are also reported. Although AWQ focuses on quantization, it complements the notion of NAS or pruning by providing a targeted approach to per-channel precision adjustments.

\subsubsection{AutoDistill: Bayesian NAS + Distillation \cite{auto-distill}}
\label{subsec:autodistill}

\begin{figure}[!ht]
    \centering
    \includegraphics[width=0.7\linewidth]{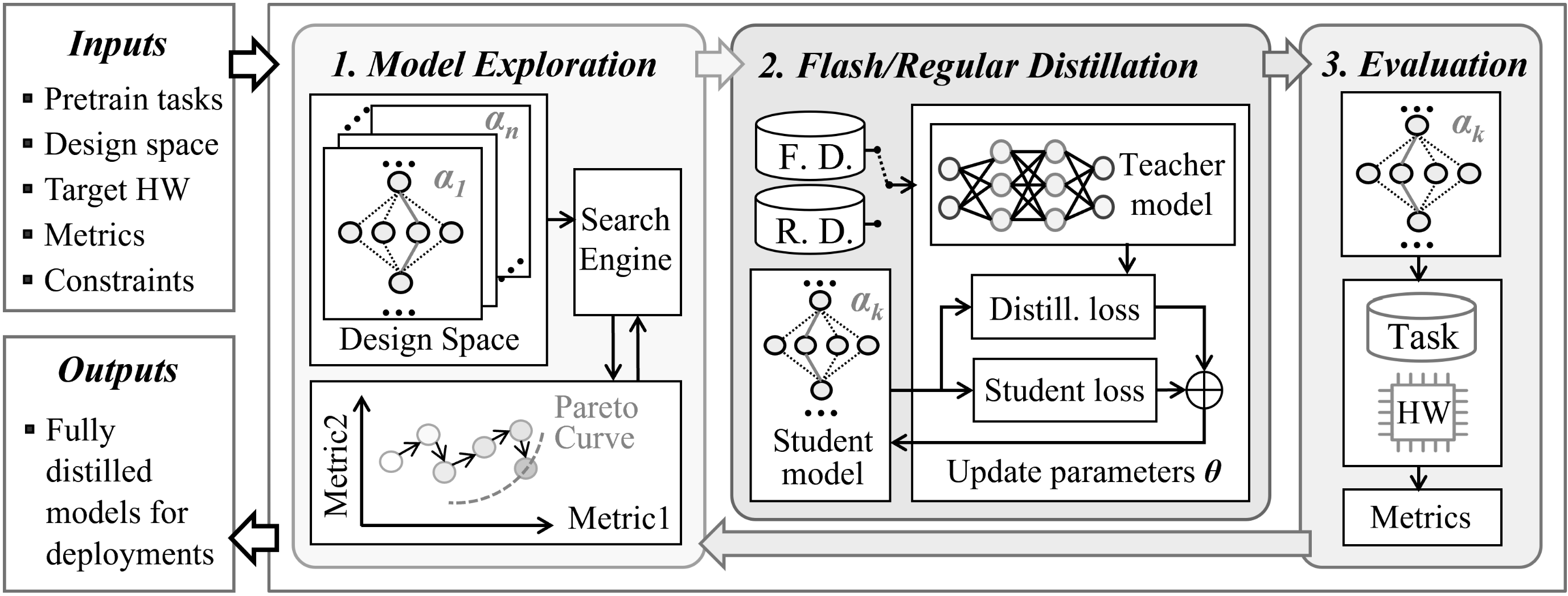}
    \vspace{-10pt}
    \caption{\textbf{AutoDistill} pipeline \cite{auto-distill}. The system repeatedly samples architectures (driven by Bayesian Optimization), does a quick ``flash'' distillation, and measures hardware metrics.}
    \label{autodistill_framework}
\end{figure}

\textbf{AutoDistill} unifies NAS, teacher-student distillation, and hardware-aware objectives. The pipeline, see Fig. \ref{autodistill_framework}:
\begin{enumerate}
    \item \textbf{Input Model and Constraints}: Start with a large, pre-trained model (the teacher), target hardware constraints (e.g., max memory, latency), and a discrete search space (24-layer stack of cells \cite{sun2020mobilebertcompacttaskagnosticbert}).
    \item \textbf{Bayesian Optimization (BO)}: Use Google Vizier \cite{golovin2017} to select candidate architectures for the student. A \emph{flash distillation} (early stopping) approach provides a fast proxy for the final accuracy after knowledge distillation, mitigating the cost of full retraining.
    \item \textbf{Distillation Loss}: Combine multi-head attention (MHA) loss, feature map (FM) loss, and logit loss \cite{sun2020mobilebertcompacttaskagnosticbert} to train the student. Mismatched teacher-student layers are handled via an intermediate fully-connected layer.
    \item \textbf{Iterative Search}: Candidate architectures are evaluated on hardware for latency, FLOPs, memory usage, etc. These measurements inform the BO’s acquisition function for the next iteration.
    \item \textbf{Full Training}: Promising architectures undergo complete distillation (200 epochs) for final accuracy. The end result is a smaller model that meets hardware constraints and approximates the teacher’s performance.
\end{enumerate}

By combining Bayesian NAS and knowledge distillation in a single loop, AutoDistill exemplifies how compression and architecture design can be simultaneously optimized for edge deployment.

\begin{figure}[!ht]
    \centering
    \includegraphics[width=0.55\linewidth]{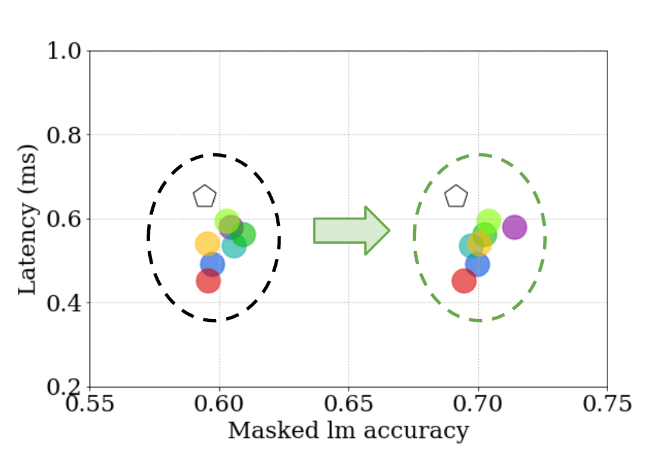}
    \caption{Flash Distillation correlates strongly with Full Distillation \cite{auto-distill}. Each point represents a candidate architecture.}
    \label{autodistill_flashdistill}
\end{figure}

\paragraph{Flash Distillation Correlation}
An intriguing component of AutoDistill is \textbf{Flash Distillation}, which uses only $\sim5\%$ of the full training steps to predict final performance. Fig.~\ref{autodistill_flashdistill} shows a high correlation between flash-distilled accuracy and fully-distilled accuracy, validating the method as a surrogate.

\subsection{Summary of Integrated Methods}

Overall, these case studies demonstrate how combining \textbf{model compression}, \textbf{NAS + HPO}, and \textbf{hardware constraints} can lead to more specialized, efficient, and robust models for edge devices. The synergy arises because each approach addresses different stages of the pipeline:

\begin{itemize}
    \item \emph{Compression} (pruning, quantization, distillation) reduces model size and complexity.
    \item \emph{NAS + HPO} explores architectural variations to find topologies inherently suited to resource constraints.
    \item \emph{Hardware-Aware Constraints} (LUT-based or direct measurement) ensure feasibility and speed on real devices.
\end{itemize}

By leveraging established benchmarks like NAS-Bench-101/201/301 and integrated frameworks such as APQ, DARTS, AWQ, and AutoDistill, researchers can systematically evaluate and refine AI models for edge deployment.

\section{Conclusion and Discussion}

Our comprehensive survey of the AI model optimization landscape aimed to identify a set of effective approaches that could inform the development of a scalable, generalized framework for optimizing AI models - one that is both model-agnostic and platform-independent. This survey not only sheds light on the  complexities of a generalized model optimization but also paves the way for formulating a mathematically tractable problem formulation that can facilitate inference acceleration across diverse AI models deployed over resource-constrained edge computing platforms.
We identified three broad methods for inference acceleration. The first method is model compression, composed of pruning, quantization, tensor decomposition, and distillation. These methods start with a large, high-performing model and find ways to achieve comparable performance with less resources. The second method is Neural Architecture Search, which searches for the optimal architecture for a given task and sometimes hardware. The third method of inference acceleration is choice of compiler; at deployment time, how the model packaged and and sent for computation. This is rapidly changing given the proliferation of new hardware architectures.

While the work in distillation has had notable victories, we note room for more fundamental work comparing loss topologies for different distillation formulations. As a practical matter, a researcher needs to estimate how much data and training time are needed to fairly evaluate a model. A comparison of this kind on a fixed task, dataset, and architecture but varying the distillation loss formulation may give the researcher an intuition for how different formulations may affect the topology. What do different distillation formulations do to the loss topology? Are there more minima? Are they lower overall? Are there more paths to the same minima? Researchers may need to know how different distillations may be suited for specific tasks. If a topology is jagged, current evidence suggests it could benefit from self-distillation which regularizes and softens the landscape.\cite{self_distillation_regularization} What other experiments and conclusions can we draw like this?

Neural Architecture Search is well studied by the AutoML community, but research frontiers still abound. Of particular interest are the Program Synthesis approaches that search for hierarchical motifs and compose new architectures in terms of those motifs, fighting the combinatorial explosion in the search space of large, complex architectures. While still an unsolved problem, advances in Neurosymbolic AI may advance the field dramatically in the near future, building custom model architectures while finding hardware-aware motifs.

\subsection{Discussion and Future Directions}\label{sec:discussion}

In this survey, we examined a range of strategies for optimizing AI models under the stringent resource constraints often found in edge computing environments. Three major categories of inference acceleration methods emerged:

\begin{enumerate}
    \item \textbf{Model Compression}: Techniques such as pruning, quantization, tensor decomposition, and distillation compress large, over-parameterized models to more manageable forms. These methods aim to preserve near state-of-the-art performance while significantly reducing memory, compute, and energy overhead.
    
    \item \textbf{Neural Architecture Search (NAS)}: NAS systematically explores architectural variations to identify designs inherently suited to the performance, latency, and memory requirements of a given task or hardware. When combined with hyperparameter optimization (HPO), it provides a comprehensive strategy for discovering configurations that exploit model redundancy and meet specific deployment constraints.
    
    \item \textbf{Compiler and Deployment Frameworks}: Specialized toolchains (e.g., TVM, TensorRT, OpenVINO) optimize model graphs at the implementation level, leveraging hardware-specific features (such as operator fusion or sparse kernels) to further accelerate inference.
\end{enumerate}

\noindent
\textbf{Challenges and Open Questions in Distillation.}  
Although knowledge distillation has yielded important real-world benefits-particularly for large language models-its theoretical underpinnings remain partially unexplored. For instance, self-distillation appears to regularize the training process \cite{self_distillation_regularization}, but there is limited understanding of why certain distillation objectives might induce flatter minima or better generalization. Furthermore, mismatches between teacher and student architectures raise questions about how to measure “distance” in latent feature space, how to handle domain gaps (e.g., cross-modal transfers), and how much data or training time is required to achieve a robust student model. Deeper investigations of the \emph{loss landscapes} shaped by distinct distillation formulations could clarify which methods are most appropriate for particular tasks or modalities.

\noindent
\textbf{Frontiers in NAS.}
While NAS has seen considerable progress in the AutoML community, several frontiers remain relatively unexplored:

\begin{itemize}
    \item \emph{Hierarchical / Program-Synthesis Approaches}: Searching for high-level motifs or symbolic expressions within vast design spaces can mitigate combinatorial blow-ups, especially for very deep networks. Advances in neurosymbolic AI may one day automate hardware-specific or domain-specific architecture discovery with minimal manual intervention.
    
    \item \emph{Scalability to Very Large Models}: Applying NAS to large language models (LLMs) remains computationally daunting; supernetwork or hypernetwork-based approaches face scalability hurdles with billions of parameters. Hybrid techniques (e.g., partial weight sharing) or approximate evaluation methods may be necessary to push NAS further into the LLM realm.
\end{itemize}

\noindent
\textbf{Sparse Literature on Pre-Training Pruning.}  
Despite the potential of pruning \emph{during} or \emph{before} large-scale training—an approach that could save considerable compute costs—existing literature is sparse for foundation-scale models. The expense of even partially training a massive model often deters researchers. More efficient or incremental pre-training pruning algorithms could enable scaling down LLMs at earlier stages, unlocking substantial resource savings.

\noindent
\textbf{Toward a Unified, Model-Agnostic Framework.}  
Collectively, these insights suggest a path toward a \emph{unified optimization pipeline} for edge AI, one that integrates:

\begin{itemize}
    \item \emph{Adaptive Compression} (e.g., distillation, quantization, pruning) guided by data- or task-specific requirements;
    \item \emph{Neural Architecture Search} that systematically navigates both macro- and micro-architecture choices (potentially via hierarchical or symbolic encodings);
    \item \emph{Hardware-Aware Compilation} ensuring that any discovered design truly yields speedups and memory savings in deployment.
\end{itemize}

A central challenge remains formalizing this pipeline as a tractable multi-objective optimization problem, balancing metrics like accuracy, latency, energy, and model size under varying device constraints.

\noindent
\textbf{Concluding Remarks.}
Although our survey did not explicitly focus on automated multi-modal architecture design, many of the discussed techniques-particularly cross-modal distillation and hardware-aware NAS-could be extended to multi-modal tasks. With deep learning models continuing to grow in complexity and scale, the convergence of model compression, NAS, and advanced compiler optimizations promises significant gains in efficiency and deployability. Bridging theoretical understanding (e.g., loss-topology analyses in distillation or hierarchical motif synthesis in NAS) with the practical realities of modern hardware stands as an exciting frontier. We hope this survey spurs further research and collaboration across these disciplines, ultimately enabling AI systems that are both powerful and feasible in real-world, resource-constrained deployments.

\section*{Acknowledgements}

This work was supported by the DEVCOM Army Research Laboratory under Cooperative Agreement No. W911NF2420176. 

\bibliographystyle{unsrtnat}
\bibliography{references} 

\end{document}